# SMMT: Siamese Motion Mamba with Self-attention for Thermal Infrared Target Tracking

Shang Zhang[1,2,3], Huanbin Zhang[1,2,3], Yujie Cui[1,2,3], Ruoyan Xiong[1,2,3], Dali Feng[4*], and Cen He[4]

[1] Hubei Key Laboratory of Intelligent Vision Based Monitoring for Hydroelectric Engineering, China Three Gorges University, Yichang 443002, China
[2] Hubei Province Engineering Technology Research Center for Construction Quality Testing Equipment, China Three Gorges University, Yichang 443002, China
[3] College of Computer and Information Technology, China Three Gorges University, Hubei, Yichang, 443002, China
[4] Public Health Department, The Second People's Hospital of Three Gorges University, Yichang 443000, China
Zhanghb13@ctgu.edu.cn

**Abstract.** Thermal infrared (TIR) object tracking often suffers from challenges such as target occlusion, motion blur, and background clutter, which significantly degrade the performance of trackers. To address these issues, this paper proposes a novel Siamese Motion Mamba Tracker (SMMT), which integrates a bidirectional state-space model and a self-attention mechanism. Specifically, we introduce the Motion Mamba module into the Siamese architecture to extract motion features and recover overlooked edge details using bidirectional modeling and self-attention. We propose a Siamese parameter-sharing strategy that allows certain convolutional layers to share weights. This approach reduces computational redundancy while preserving strong feature representation. In addition, we design a motion edge-aware regression loss to improve tracking accuracy, especially for motion-blurred targets. We conduct extensive experiments on LSOTB-TIR, PTB-TIR, VOT-TIR2015, and VOT-TIR2017. The results show that SMMT achieves superior performance in TIR target tracking.

## 1  Introduction

Thermal Infrared (TIR) target tracking is an important research focus in the field of computer vision, particularly suitable for scenarios where traditional optical tracking methods struggle to perform effectively. While visible-light-based tracking has been extensively studied and benefits from rich features and detailed textures, it faces limitations in challenging environments. Adverse weather conditions such as fog, rain, and low light can significantly degrade the performance of visible-light tracking. In contrast, TIR tracking captures the thermal radiation naturally emitted by objects, enabling robust performance regardless of lighting conditions. This robustness makes TIR tracking highly valuable for applications such as assisted driving, environmental monitoring, and industrial automation. Consequently, developing more efficient, accurate, and robust TIR tracking methods for real-world deployment has become an increasingly urgent and practical demand.

Siamese network-based trackers have demonstrated high accuracy and robustness, making them increasingly popular in the field of visual object tracking. Bertinetto et al. introduced SiamFC [1], a fully convolutional Siamese network trained end-to-end, which tracks objects by computing the similarity between input features—significantly improving tracking performance. Building on this, Li et al. proposed SiamRPN++ [2], which incorporates a ResNet-50 backbone to enhance feature extraction and achieve further performance gains. Since then, numerous Siamese trackers have been developed, continually advancing tracking performance in complex and challenging scenarios.

However, TIR images present inherent challenges such as low resolution, a low signal-to-noise ratio, and sparse detail, which significantly increase the difficulty of distinguishing targets from the background in TIR tracking [11]. Moreover, factors like object occlusion and motion blur caused by fast-moving objects further complicate the tracking process. Various efforts have been made to improve motion modeling in tracking. For instance, the Kalman filter method proposed by Cao et al. [4] is widely used, but its accuracy is limited. Yao et al. [5] introduced a learning-based motion modeling technique, but it overlooks global motion modeling. In TIR tracking, the absence of effective global motion modeling significantly restricts tracking accuracy. To address this challenge and ensure robust tracking in complex thermal environments, we propose a novel motion modeling framework. As shown in Fig. 1, the proposed tracker is visually compared with several state-of-the-art trackers on the thermal infrared benchmark.

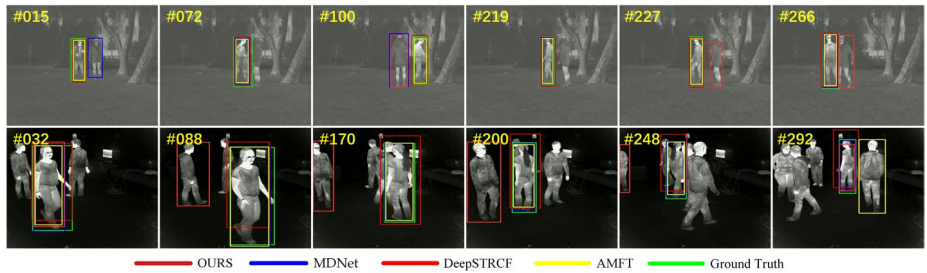

**Fig. 1.** The visualization shows the tracking results of the four trackers on the thermal infrared benchmarks, with blue, red, yellow, and rose boxes representing MDNet, DeepSTRCF, AMFT, and the proposed SMMT, respectively.

To address these challenges, we propose SMMT tracker, designed for fast and accurate motion modeling and tracking. Unlike traditional Siamese trackers, SMMT incorporates a bidirectional state-space model along with a Siamese parameter-sharing strategy, significantly reducing the computational overhead typically associated with motion modeling. By introducing the Siamese Motion Mamba module into our framework,

SMMT greatly enhances the stability and precision of motion modeling. In addition, we introduce a motion edge aware loss, which effectively improves the detection of motion-blurred targets, thereby boosting tracking performance in complex scenarios.

The main contributions of this paper are summarized as follows:

- We propose a novel Siamese motion Mamba module, which combines vertical and horizontal state-space models with self-attention blocks, improving both tracking accuracy and computational efficiency.
- We introduce a Siamese parameter-sharing strategy to reduce computational redundancy within the Siamese motion Mamba module, maintaining strong feature representation while enhancing efficiency.
- We design a motion edge-aware regression loss, which applies targeted supervision to predicted motion boundaries and edge details, significantly enhancing the detection of motion-blurred targets.
- Extensive experiments conducted on four benchmarks demonstrate the strong performance and robustness of SMMT in TIR tracking.

## 2    Related Work

### 2.1    Siamese network-based Target Tracking

Siamese network-based trackers have attracted significant attention in the field of target tracking due to their strong feature representation capabilities. These trackers typically consist of three core components: feature extraction, cross-correlation-based feature fusion, and a prediction head. The process begins by extracting features from the input images, fusing them via cross-correlation, and generating a response map to locate the target. A pioneering work in this area, SiamFC [1], utilizes a fully convolutional Siamese network for single-target tracking. It combines feature maps from the template and search branches using a cross-correlation layer, which convolves the search region with template features to produce a response map.

Building on SiamFC, Li et al. proposed SiamRPN [2], incorporating a region proposal network (RPN) and a regression branch to achieve a better balance between accuracy and speed. SiamRPN++ [3] further optimized the architecture by enhancing feature extraction and introducing a hierarchical aggregation strategy to improve robustness. SiamAPN++ [6] introduced an attention enhancement module to boost tracking performance using attention mechanisms. Inspired by SiamBAN [7], proposed by Chen et al., many subsequent trackers adopted attention mechanisms to focus on key target features and improve robustness in complex scenarios. For instance, Yu et al. combined deformable self-attention with cross-attention mechanisms to introduce SiamAttn[8], further advancing tracking accuracy. Consequently, Siamese trackers continue to evolve and have become the dominant method in TIR target tracking [13].

## 2.2   Motion Modeling in TIR Tracking

Existing Thermal Infrared (TIR) trackers often employ a hybrid strategy that combines both detection and tracking. These strategies generally fall into two categories: position-based [9] and appearance-based approaches [10]. Position-based methods focus on modeling the target's motion to predict its location in future frames [11], [14], while appearance-based trackers extract visual features of the target to measure similarity across frames for tracking [15]. However, TIR images present inherent challenges, such as low signal-to-noise ratios and limited contrast, making it more difficult to extract meaningful information compared to visible spectrum images. Consequently, we focus on improving motion modeling as the key solution to address these challenges and enhance TIR tracking performance.

In TIR tracking tasks, the Kalman filter and its variants are among the most commonly used motion modeling methods. However, these approaches typically require a priori assumptions about the target's motion patterns and their probability distributions, and they do not incorporate learnable parameters. As a result, their performance often degrades in complex scenarios involving object deformation or rapid motion. To overcome these limitations, recent studies have introduced learnable neural network-based motion modeling methods. These approaches generally outperform traditional rule-based methods because they can adapt to varying motion patterns through learning. However, a common drawback of these neural network methods is the repeated extraction of features, which leads to redundant computation and reduced efficiency.

## 3   Methodology

### 3.1   Overview of SMMT

To address the degradation in tracking accuracy caused by occlusion and motion blur, we designed a Siamese Motion Mamba (SMM) module equipped with an adaptive scale selection mechanism [17]. To enhance computational efficiency, we incorporate a Siamese parameter-sharing strategy, which helps reduce computational redundancy. Additionally, to better capture fine-grained target details, we introduce a motion edge-aware regression loss, which provides supervision for both the classification and regression branches. The backbone network is responsible for extracting detection features from each video frame. It produces feature maps at three different scales, 1/4, 1/8, and 1/16 of the original image size, considering the inherent limitations of TIR images. The SMM module then extracts motion features from these feature maps using the adaptive scale selection mechanism. These features are fused via cross-correlation and passed to the region proposal network (RPN) to predict the location of the target.

### 3.2   Siamese Motion Mamba module

To address the limitations in motion modeling capability and the large parameter size commonly found in Siamese network-based trackers, we propose the Siamese Motion

Mamba (SMM) module, a lightweight and efficient solution for motion modeling. This module extracts motion features across three spatial scales, beginning feature fusion at the lowest resolution and ultimately producing a motion feature map at 1/8 the size of the original image [17]. Given the inherent challenges of TIR images, such as low resolution and limited detail, we introduce an adaptive scale weighting mechanism based on the target size. Specifically, two thresholds, $T_{small}$ and $T_{large}$, are defined:

- If the target region size $S_r$ is smaller than $T_{small}$, the feature weight at the 1/16 scale is emphasized.
- If the target size $S_r$ exceeds $T_{large}$, the feature weight at the 1/4 scale is enhanced.
- When the target size falls between these thresholds, the feature weight at the 1/8 scale is enhanced by default.

This adaptive strategy allows the tracker to better handle varying target sizes under the constraints of TIR imaging.

As illustrated in Fig.2, each SMM module is composed of two branches: the Horizontal State-Space (HSS) model and the Vertical State-Space (VSS) model. These two branches of scan feature maps in horizontal and vertical directions, respectively [16]. For a feature map with height $H$, width $W$, and $C$ channels, the HSS branch performs sequential scans across all rows, starting from the first column of each row and continuing to the last column, resulting in $H$ total scans. Similarly, the VSS branch performs $W$ scans, processing the feature map column-wise, analogous to the HSS operation.

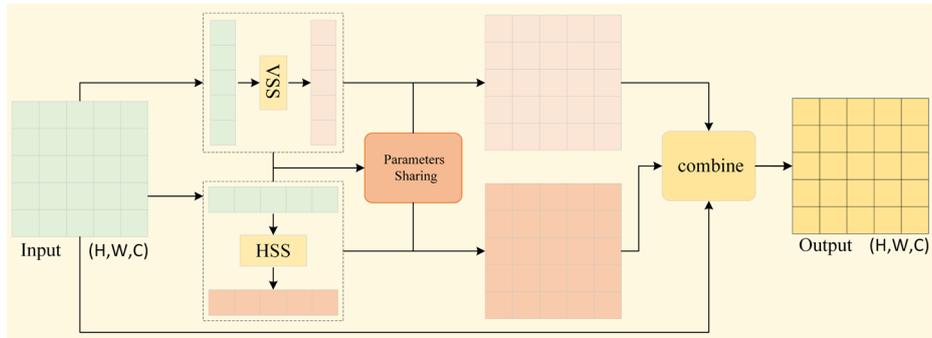

**Fig. 2.** The architecture of the proposed Siamese Motion Mamba. The vertical state space model (VSS) and the horizontal state space model (HSS) are used to scan the feature map in two directions.

The SMM module extracts motion features from detection features of both the previous and next frame images. However, this process can be sensitive to target occlusion, potentially leading to tracking failure. To mitigate this issue, and inspired by the tracker proposed by Kautz et al. [18], we integrate self-attention blocks into the final layer of the SMM module. By combining the SMM module with a Transformer-style self-attention mechanism, we seek to enhance the temporal stability of the motion features.

The proposed SMMT tracker employs a standard multi-head self-attention mechanism, formulated as follows:

$$Attention(Q, K, V) = softmax\left(\frac{QK^T}{\sqrt{d_h}}\right)V \quad (1)$$

Here, $Q, K, V$ represent the query, key, and value matrices, respectively, and $d_h$ is the dimensionality of each attention head. The softmax operation assigns higher weights to more relevant motion features, allowing the model to better capture and preserve target information even under occlusion.

### 3.3 Siamese Parameter-sharing Strategy

Although the Siamese Motion Mamba module enables fast motion modeling through its Vertical State Space (VSS) and Horizontal State Space (HSS) components, its bidirectional scanning mechanism introduces redundant parameter usage, significantly increasing computational overhead. To address this issue and enhance computational efficiency, we propose a Siamese Parameter-sharing Strategy (SPS). This strategy reduces redundancy by allowing HSS and VSS to share parameters in specific convolutional layers, thereby improving processing speed without compromising performance. The core idea is to maintain a balance between reducing the number of parameters and preserving the feature representation capability. Designing an effective SPS is a key focus of our work, as it directly influences this trade-off.

Inspired by the backbone network splitting strategy in [19], we divide the ResNet-50 backbone into two functional parts:

- The first part (Layers 0 to 2) follows the standard ResNet-50 structure and is used to extract fundamental features from the input.
- The second part (Layers 3 and 4) incorporates the SPS, enabling HSS and VSS to share convolutional parameters when scanning the feature maps.

This design significantly reduces redundant computation during motion modeling, resulting in a more efficient and lightweight tracking framework.

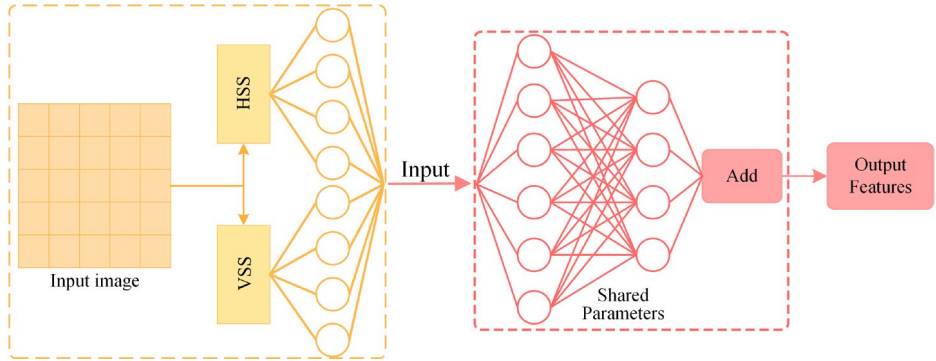

**Fig. 3.** The architecture of the proposed Siamese parameter-sharing strategy.

We denote the input images of the template region and search region as $I_Z$ and $I_X$, respectively. The corresponding low-level features extracted from these regions are represented as $F_{base}^Z$ and $F_{base}^X$. The VSS and HSS are denoted by $\varphi_V$ and $\varphi_H$, respectively. These components are responsible for scanning and extracting features along the vertical and horizontal directions. The final output features of the network are represented as $F \in \{F_V^Z, F_H^Z, F_V^X \text{ and } F_H^X\}$, and are defined as follows:

$$\begin{cases} F_V^Z = \varphi_{SH}(\varphi_V(F_{base}^Z)) \\ F_H^Z = \varphi_{SH}(\varphi_H(F_{base}^Z)) \\ F_V^X = \varphi_{SH}(\varphi_V(F_{base}^X)) \\ F_H^X = \varphi_{SH}(\varphi_H(F_{base}^X)) \end{cases} \quad (2)$$

Here, $\varphi_{SH}$ denotes the Siamese parameter-sharing strategy. Specifically, $F_V^Z$ and $F_V^X$ are the features extracted by VSS from the template and search regions. $F_H^Z$ and $F_H^X$ are the features extracted by HSS from the template and search regions.

### 3.4    Loss Function of SMMT

When a target is occluded by other objects during motion, its features may become partially or entirely lost. In addition, camera rotations can result in rapid displacement of the object within the image, leading to motion blur. These challenges significantly hinder tracking accuracy. Even brief interruptions in target tracking can severely degrade overall performance. Given the frequency of such cases in TIR datasets, we introduce a Motion Edge-aware Regression Loss Function specifically designed to address these issues and enhance tracking robustness.

The proposed loss function is composed of three components: regression loss, motion edge loss, and fine-grained feature loss, formulated as:

$$L_{RFME} = \lambda_{reg}L_{reg} + \lambda_{meg}L_{meg} + \lambda_{fgl}L_{fgl} \quad (3)$$

where $\lambda_{reg}$, $\lambda_m$ and $\lambda_f$ are the weighting coefficients for each component.

**Regression Loss $L_{reg}$.** The baseline tracker SiamRPN++ utilizes a Region Proposal Network (RPN) for bounding box regression. Thus, incorporating a regression loss is essential for accurate target localization. The regression loss is calculated using a weighted L1 loss, combining IoU and coordinate-based error:

$$L_{reg} = \frac{1}{N}\sum_{j=1}^{N}\left[\alpha L_{\text{IoU}}(b_j, \hat{b}_j) + \beta L_1(t_j, \hat{t}_j)\right] \quad (4)$$

where $N$ is the number of positive samples, $b_j$ and $\hat{b}_j$ are the ground truth and predicted bounding boxes for the $j$-th sample, respectively. $t_j$ and $\hat{t}_j$ are the ground truth and predicted coordinates. $\alpha$ and $\beta$ are loss weights.

**Motion Edge Loss $L_{meg}$.** Due to the low resolution and limited texture in TIR images, we propose motion edge loss to preserve critical edge information. This loss helps refine boundaries of motion-blurred targets:

$$L_{meg} = \sum_{i=1}^{N} \left( u \parallel \nabla I_i - \nabla \hat{I}_i \parallel_2^2 + v \parallel M_i \odot (\nabla I_i - \nabla \hat{I}_i) \parallel_1 \right) \quad (5)$$

where $\nabla I_i$ and $\nabla \hat{I}_i$ are the gradients of the ground truth and predicted images, $M_i$ is the binary mask for the $i$-th image, and $u, v$ are weight coefficients.

**Fine-Grained Feature Loss $L_{fgl}$.** To counteract the lack of detail and environmental noise in TIR imagery, we propose a Fine-Grained Feature Loss, which enhances local feature sensitivity and strengthens tracking robustness:

$$L_{fgl} = (\lambda_1 + \gamma S) \log(1 + E_F) + \lambda_2 \log(1 + E_W) \quad (6)$$

where $\lambda_1$ and $\lambda_2$ are constant weights, $\gamma$ adjusts the global feature error based on scale $S$, and $E_F$, $E_W$ represent global and local feature alignment errors, defined as:

$$\begin{cases} E_F = -\sum_i p_i \log(q_i) \\ E_W = \sum_{i=1}^{N} \left\| W_i \odot (F_i - \hat{F}_i) \right\|_1 \end{cases} \quad (7)$$

where $E_F$ measures global feature matching via cross-entropy between distributions $p$ and $q$, while $E_W$ captures local feature alignment using $L1$ with a weighting $W_i$.

## 4    Experiment

In this section, we evaluate the performance of SMMT across four datasets: LSOTB-TIR [21], PTB-TIR [22], VOT-TIR 2015 [23], and VOT-TIR 2017 [24]. We first describe the experimental setup, including the evaluation metrics and datasets used. Then, we examine several key model parameters and analyze the robustness of the proposed method. Finally, SMMT was compared with several mainstream trackers, and then the visual tracking results were shown.

### 4.1    Implementation Details

**Datasets and Evaluation Criteria.** The VOT-TIR2015 dataset was the first benchmark designed specifically for evaluating short-term thermal infrared (TIR) target tracking methods. It contains 20 TIR tracking sequences across eight categories of tracking targets, with an average sequence length of 563 frames. Building upon this, the VOT-TIR2017 dataset includes 25 TIR tracking sequences, with an average length of 740 frames. This larger dataset presents a more challenging and diverse test for tracking algorithms. The LSOTB-TIR dataset stands as one of the largest and most diverse

TIR tracking datasets available today, featuring over 1,400 TIR video sequences, more than 600,000 frames, and annotations for over 730,000 bounding boxes, making it the largest dataset for long-term TIR tracking. Lastly, the PTB-TIR dataset focuses on TIR pedestrian tracking, comprising 60 sequences with a total of over 30,000 frames. This dataset provides a valuable resource for assessing pedestrian-specific tracking methods.

Following [21] and [22], we use center location error (Precision) and overlap score (Success) as evaluation metrics for the PTB-TIR benchmark. For the LSOTB-TIR benchmark, we additionally incorporate Normalized Precision (NP) as an evaluation metric. Following [23] and [24], we utilize precision, robustness, and expected average overlap (EAO) as the primary evaluation metrics for the VOT-TIR2015 and VOT-TIR2017 benchmarks.

**Experimental Platform.** The proposed SMMT is implemented in Python 3.7.1, and all experiments are conducted on a computer running Ubuntu 20.04, featuring an Intel i5-12400H CPU, 32GB of RAM and an NVIDIA RTX 3060 GPU.

### 4.2  Ablation Experiment

**Ablation Study of Methods.** In this section, we conduct several comparative experiments to evaluate the effectiveness of each component of the proposed tracker. First, we compare Siam-SMM with SiamRPN++ to demonstrate the contribution of the Siamese Motion Mamba module to the overall tracker performance. Next, we compare Siam-SPS with SiamRPN++ to validate the reliability of the proposed Siamese parameter-sharing strategy. Finally, we integrate the SMM, Siamese parameter-sharing strategy, and motion edge-aware loss into SiamRPN++ and compare the results to demonstrate the effectiveness of these components.

The ablation experimental results on LSOTB-TIR are shown in Table 1, with SiamRPN++ as the baseline model. The experiments demonstrate that the proposed methods contribute to improving tracking performance. First, compared to SiamRPN++, Siam-SMM achieves improvements of 3.8 in precision and 3.2 in success rate, indicating that the Siamese Motion Mamba module enhances the focus on target details, thus improving tracking accuracy. After incorporating the Siamese parameter-sharing strategy, Siam-RFME achieves improvements of 7.2 in precision and 6.8 in success rate. This demonstrates that the Siamese parameter-sharing strategy helps reduce computational redundancy, thereby enhancing tracking precision to some extent.

Table 1. Ablation analysis on LSOTB-TIR benchmark.

| Model | Component | | | Precision (P/%) | Norm. Pre. (NP/%) | Success (S/%) |
| --- | --- | --- | --- | --- | --- | --- |
| | SMM | SPS | RFME | | | |
| SiamRPN++ (base) | | | | 74.0 | 69.2 | 55.4 |
| Siam-SMM | √ | | | 77.8 | 71.9 | 58.6 |
| Siam-SPS | | √ | | 76.2 | 70.8 | 59.0 |
| Siam-RFME | √ | √ | | 81.2 | 71.4 | 62.2 |
| SMMT (proposed) | √ | √ | √ | **85.4** | **74.6** | **62.8** |

**Ablation Study of Parameters.** To further examine the influence of the coefficients in $L_{\text{reg}}$ on tracker performance, we conduct ablation experiments on the IoU loss coefficient $\alpha$ and the L1 loss coefficient $\beta$. This experiment aims to explore what values of α and β can make the best precision and success rate of the tracker.

Fig. 4 and 5 present the precision and success rate of the tracker on the PTB-TIR benchmark for different values of $\alpha$ and $\beta$, respectively. The results indicate that when $\alpha$ is set to 0.80 and $\beta$ is set to 0.50, the tracker achieves the highest precision and success rate. Therefore, in this work, $\alpha$ is set to 0.8 and $\beta$ is set to 0.5.

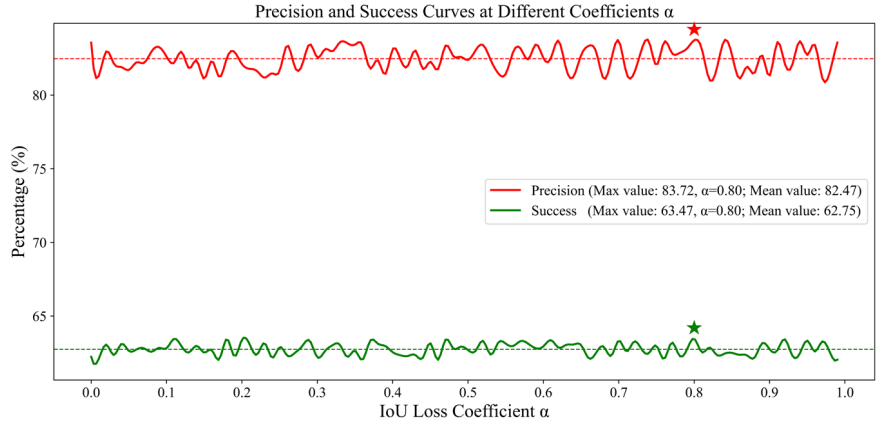

**Fig. 4.** Precision and success rate of the tracker with different loss coefficient $\alpha$.

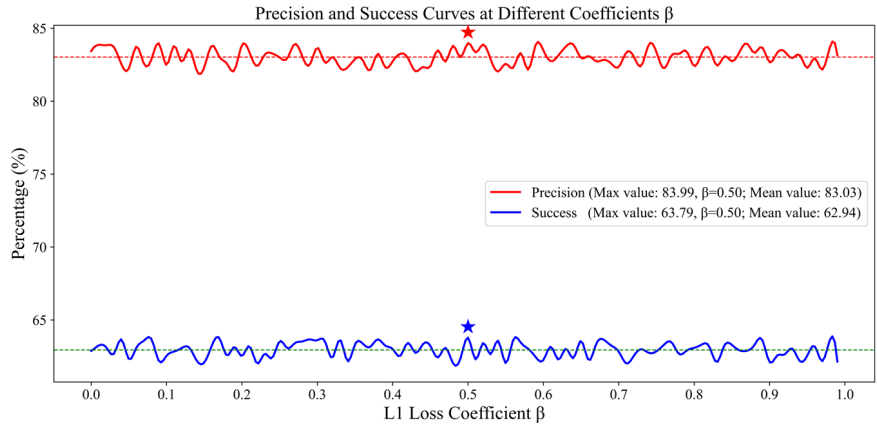

**Fig. 5.** Precision and success rate of the tracker with different loss coefficient $\beta$.

**Performance Comparison with State-of-the-arts**

To evaluate the proposed SMMT algorithm, we compare it with several state-of-the-art trackers, including Siamese network based trackers like SiamRPN++ [3], SiamFC [1], TADT [25], CFNet [26], and MLSSNet [27]; Transformer-based trackers such as VITAL [28]; deep learning-based trackers like DeepSTRCF [29], MDNet [30], HSSNet [31], and ATOM [32]; as well as correlation filter-based trackers such as ECO-stir [33], ECO-HC [34], Staple [35], MCCT [36], DSST [37], KCF [38], and HCF [39]. The experimental results on the LSOTB-TIR, PTB-TIR, VOT-TIR2015, and VOT-TIR2017 benchmarks are shown in Fig. 4, Fig. 5, Table 2, Table 3, and Fig. 6, respectively.

**Results on PTB-TIR.** As shown in Fig. 6 (a) and (b), our tracker achieves the highest precision and success rate on PTB-TIR, reaching 83.9% and 63.6%, respectively. Compared to MDNet [29], which incorporates deep learning, our tracker improves precision by 3.4 percentage points and increases the success rate by 5.0 percentage points. In comparison with the correlation filter-based Staple tracker [34], the proposed tracker demonstrates improvements of 9.6 percentage points in precision and 9.0 percentage points in success rate.

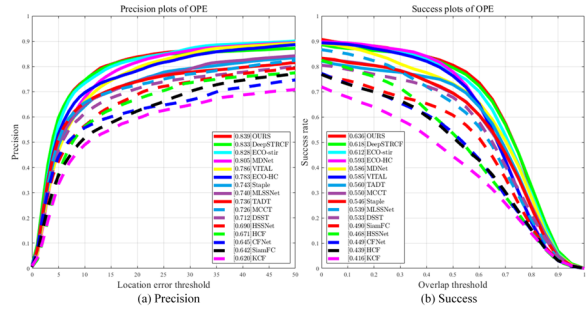

**Fig. 6.** Performance comparison results on PTB-TIR benchmark.

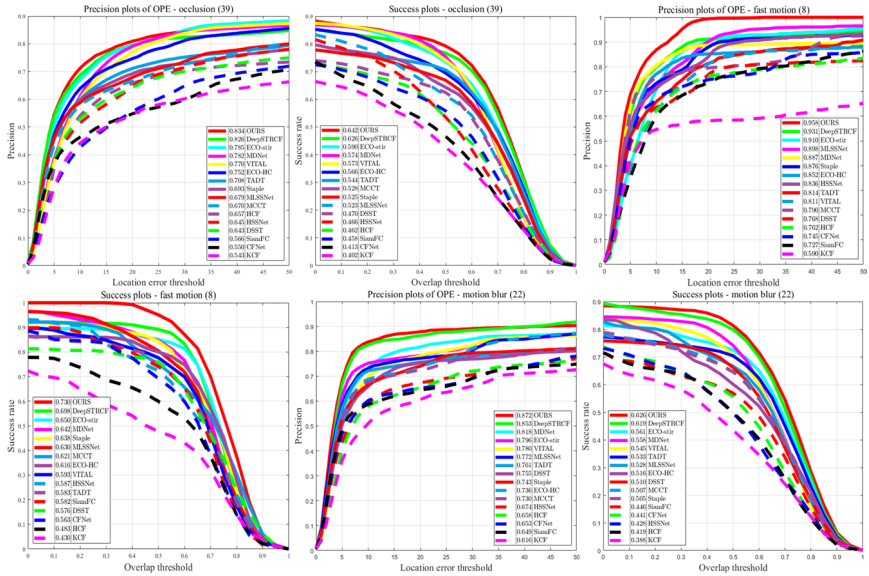

**Fig. 7.** Performance comparison results on the PTB-TIR benchmark in three scenarios (scenarios including occlusion, fast motion, motion blur).

To evaluate the effectiveness of SMMT, we compare it with several state-of-the-art trackers in three challenging scenarios from the PTB-TIR benchmark: occlusion, fast motion, and motion blur. As shown in Fig. 7, our proposed tracker outperforms all others, achieving the highest precision and success rate in all three scenarios. The excellent performance of SMMT can be attributed to the Siamese Motion Mamba module, which performs motion modeling for motion blurred and partially occluded targets through a bidirectional state-space model and multiple self-attention blocks. To achieve the best performance of SMMT in fast motion scenes, computational redundancy in the Siamese Motion Mamba module is minimized to the greatest extent through the Siamese parameter-sharing strategy. Moreover, the incorporation of a motion edge-aware regression loss significantly enhances tracking precision under motion blur conditions.

**Table 2.** Tracker performance on VOT-TIR2015 and VOT-TIR2017 benchmarks.

| Methods | Trackers | Year | VOT-TIR 2015 | | | VOT-TIR 2017 | | |
|---|---|---|---|---|---|---|---|---|
| | | | EAO ↑ | Acc ↑ | Rob ↓ | EAO ↑ | Acc ↑ | Rob ↓ |
| Correlation filter | SRDCF | 2015 | 0.225 | 0.62 | 3.06 | 0.197 | 0.59 | 3.84 |
| | ECO-deep | 2017 | 0.286 | 0.64 | 2.36 | 0.267 | 0.61 | 2.73 |
| | ATOM | 2019 | 0.331 | 0.65 | 2.24 | 0.290 | 0.61 | 2.43 |
| | ECO-MM | 2022 | 0.303 | 0.72 | 2.44 | 0.291 | 0.65 | 2.31 |
| | ECO_LS | 2023 | 0.319 | 0.64 | 0.82 | 0.302 | 0.55 | 0.93 |
| Transformer | TransT | 2021 | 0.287 | 0.77 | 2.75 | 0.290 | 0.71 | 0.69 |
| | DFG | 2024 | 0.329 | 0.78 | 2.41 | 0.304 | **0.74** | 2.63 |
| | CorrFormer | 2023 | 0.269 | 0.71 | **0.56** | 0.262 | 0.66 | 1.23 |
| Deep learning | DeepSTRCF | 2018 | 0.257 | 0.63 | 2.93 | 0.262 | 0.62 | 3.32 |
| | DiMP | 2019 | 0.330 | 0.69 | 2.23 | 0.328 | 0.66 | 2.38 |
| | Ocean | 2020 | 0.339 | 0.70 | 2.43 | 0.320 | 0.68 | 2.83 |
| | UDCT | 2022 | **0.420** | 0.67 | 0.88 | 0.342 | 0.66 | 0.81 |
| Siamese network | SiamFC | 2016 | 0.219 | 0.60 | 4.10 | 0.188 | 0.50 | **0.59** |
| | CFNet | 2017 | 0.282 | 0.55 | 2.82 | 0.254 | 0.52 | 3.45 |
| | DaSiamRPN | 2018 | 0.311 | 0.67 | 2.33 | 0.258 | 0.62 | 2.90 |
| | SiamRPN | 2018 | 0.267 | 0.63 | 2.53 | 0.242 | 0.60 | 3.19 |
| | TADT | 2019 | 0.234 | 0.61 | 3.33 | 0.262 | 0.60 | 3.18 |
| | SiamRPN++ | 2019 | 0.313 | 0.74 | 2.25 | 0.296 | 0.69 | 2.63 |
| | MMNet | 2020 | 0.340 | 0.61 | 2.09 | 0.320 | 0.58 | 2.90 |
| | MLSSNet | 2020 | 0.329 | 0.57 | 2.42 | 0.286 | 0.56 | 3.11 |
| | SMMT (ours) | 2025 | 0.376 | **0.79** | 1.58 | **0.345** | **0.74** | 1.95 |

**Results on VOT-TIR2015 and VOT-TIR2017.** As shown in Table 2, the proposed SMMT achieves the highest accuracy scores of 0.79 and 0.74 on the VOT-TIR2015 and VOT-TIR2017 benchmarks, respectively, surpassing previous state-of-the-art trackers such as SiamRPN++, MMNet, Ocean, and ECO-MM. SMMT achieves a score of 0.376 on VOT-TIR2015 and the best EAO score of 0.345 on VOT-TIR2017. Experimental results on vote-tir2015 and vote-tir2017 show that compared with the tracker SiamRPN, which is also based on Siamese network, the accuracy of our tracker is improved by 16% and 14.0%, respectively.

### 4.3  Visualized Comparison Results

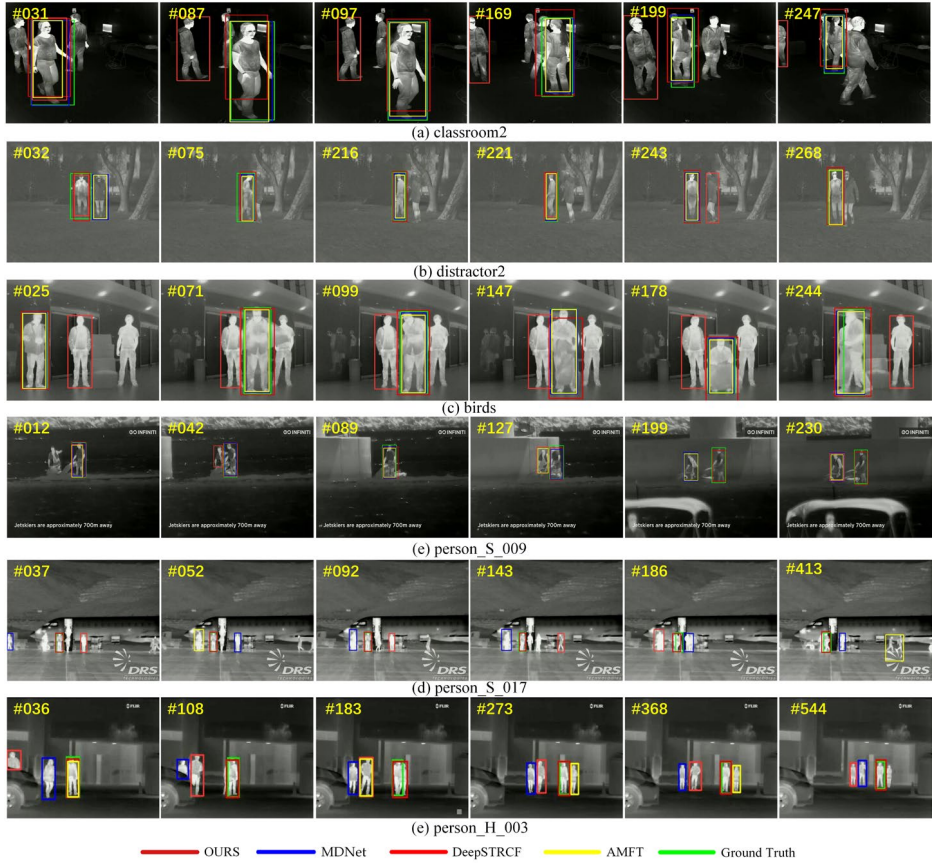

**Fig. 8.** Visualization results of the qualitative comparison experiments conducted on six challenging sequences in the PTB-TIR and LSOTB-TIR benchmarks (In order: classroom2, distractor2, birds, room1, person_S_009, person_S_017, and person_H_003)

To evaluate the tracking performance, visual tracking results of SMMT and several trackers (MDNet, DeepSTRCF, and AMFT) are presented in Fig. 8. As shown in Fig. 8, the proposed SMMT outperforms the other trackers in six tracking sequences, with the tracking bounding boxes being closest to the ground truth. Notably, DeepSTRCF and MDNet are sensitive to scenarios involving target occlusion and interference, leading to target loss during tracking. In sequences such as 'classroom2', 'birds', and 'person_S_017', which present common challenges like target occlusion and motion blur, SMMT effectively addresses these issues by utilizing the Siamese Motion Mamba module for motion modeling. Additionally, these sequences also involve challenges like thermal crossover and motion blur, where SMMT performs well, while MDNet, AMFT, and DeepSTRCF mistakenly identify certain background elements or incorrect features as the target. These results demonstrate that SMMT outperforms other TIR trackers in terms of performance.

## 5    Conclusions

In this paper, we propose a novel Siamese Motion Mamba Tracker (SMMT). This tracker is designed to address the challenges of motion blur and severe occlusion in TIR tracking. By integrating a bidirectional state-space model and a self-attention mechanism, SMMT effectively extracts motion features and recovers overlooked edge details. We introduce a Siamese parameter-sharing strategy to reduce computational redundancy in the Siamese Mamba module, improving tracking efficiency. Additionally, we design a motion edge-aware regression loss to enhance tracking accuracy, particularly for motion-blurred targets. Extensive experiments on the VOT-TIR2015, VOT-TIR2017, PTB-TIR, and LSOTB-TIR benchmarks demonstrate that SMMT outperforms state-of-the-art trackers in both precision and robustness. In future work, we aim to incorporate more efficient parameter-sharing strategies and motion mamba modules to better meet the demands of object tracking in real-world scenarios.

## References

1. Bertinetto, L., Valmadre, J., Henriques, J. F., Vedaldi, A., Torr, P. H.: Fully-convolutional siamese networks for object tracking. In: ECCV Workshops, 14, 850-865 (2016)
2. Li, B., Yan, J., Wu, W., Zhu, Z., Hu, X.: High performance visual tracking with siamese region proposal network. In: CVPR, 8971-8980 (2018)
3. Li, B., Wu, W., Wang, Q., Zhang, F., Xing, J., Yan, J.: Siamrpn++: Evolution of siamese visual tracking with very deep networks. In: ICCV, 4282-4291 (2019)
4. Cao, J.; Pang, J.; Weng, X.; Khirodkar, R.; and Kitani, K.: Observation-centric sort: Rethinking sort for robust multi-object tracking. In: CVPR, 9686-9696 (2023)
5. Yao, M.; Wang, J.; Peng, J.; Chi, M.; and Liu, C.: FOLT: Fast Multiple Object Tracking from UAV-captured Videos Based on Optical Flow. In: ICME, 3375–3383 (2023)
6. Cao, Z.; Fu, C.; Ye, J.; Li, B.; Li, Y.: SiamAPN++: Siamese attentional aggregation network for real-time UAV tracking. In: IROS, 3086–3092 (2021)
7. Chen, Z., Zhong, B., Li, G., Zhang, S., Ji, R., Tang, Z., et al.: SiamBAN: Targetaware tracking with siamese box adaptive network. In: Trans. Pattern Anal. Machine Intell, 5158–5173 (2022)
8. Yu Y, Xiong Y, et al.: Deformable Siamese attention network for visual object tracking. In: CVPR, 6728-6737 (2020)
9. Zhao, X.; Gong, D.; and Medioni, G. 2012.: Tracking using motion patterns for very crowded scenes. In ECCV, 315-328 (2012)
10. Yao T, Hu J, Zhang B, et al.: Scale and appearance variation enhanced siamese network for thermal infrared target tracking. In: Infrared Physics & Technology, 103825 (2021)
11. Shuai, B.; Berneshawi, A.; Li, X.; Modolo, D.; and Tighe, J.: SiamMOT: Siamese Multi-Object Tracking. In: IEEE Conf. Comput. Vis. Pattern Recog, 12372–12382 (2021)
12. Xiong, R., Zhang, S., Zhang, Y., & Zhang, M.: SPECTER: A Tracker with Super-Resolution Siamese Network Reconstruction and Multi-Scale Feature Extraction for Thermal Infrared Pedestrian Tracking. In: ICPRAI. 7-16 (2024)

13. Xiong, R., Zhang, S., Zou, Y., & Zhang, Y.: SRCFT: A Correlation Filter Tracker with Siamese Super-Resolution Network and Sample Reliability Awareness for Thermal Infrared Target Tracking. In: ICIC, 145-156 (2024)
14. Zhang, Y.; Sun, P.; Jiang, Y.; Yu, D.; Weng, F.; Yuan, Z.; Luo, P.; Liu, W.; and Wang, X. Bytetrack: Multiobject tracking by associating every detection box. In: ECCV, 1–21. (2022)
15. Peng, J.; Wang, C.; Wan, F.; Wu, Y.; Wang, Y.; Tai, Y.; Wang, C.; Li, J.; Huang, F.; and Fu, Y.: Chained-tracker: Chaining paired attentive regression results for end-to-end joint multiple-object detection and tracking. In ECCV, 145161 (2020)
16. Gu A, Dao T.: Mamba: Linear-time sequence modeling with selective state spaces. In: arXiv preprint arXiv, 2312.00752 (2023)
17. Yao M, Peng J, He Q, et al.: MM-Tracker: Motion Mamba with Margin Loss for UAV-platform Multiple Object Tracking. In: arXiv preprint arXiv, 2407.10485 (2024)
18. Hatamizadeh A, Kautz J.: Mambavision: A hybrid mamba-transformer vision backbone. In: arXiv preprint arXiv, 2407.08083 (2024)
19. Chan S, Du F, Tang T, et al.: Parameter sharing and multi-granularity feature learning for cross-modality person re-identification. In: Complex & Intelligent Systems, 949-962 (2024)
20. Ye M, Wang Z, Lan X, Yuen PC.: Visible thermal person re-identification via dual-constrained top-ranking. In: IJCAI 1092–1099 (2018)
21. Liu, Q., He, Z., Li, X., Zheng, Y.: PTB-TIR: A thermal infrared pedestrian tracking benchmark. IEEE Trans. Multimedia, **22**, 666-675 (2019)
22. Liu, Q., Li, X., He, Z., Li, C., Li, J., Zhou, Z., Zheng, F.: LSOTB-TIR: A large-scale high-diversity thermal infrared object tracking benchmark. In: ACM MM, 3847-3856 (2020)
23. Felsberg, M., Berg, A., Hager, G., Ahlberg, J., et al.: The thermal infrared visual object tracking VOT-TIR2015 challenge results. In: ICCV Workshops, 76-88 (2015)
24. Kristan, M., Leonardis, A., Matas, J., Felsberg, M., Pflugfelder, R., et al.: The visual object tracking vot2017 challenge results. In: ICCV Workshops, 1949–1972 (2017)
25. Li, X., Ma, C., Wu, B., et al.: Target-aware deep tracking. In: ICCV, 1369-1378 (2019)
26. Zhang, G., Li, Z., Tang, C., Li, J., Hu, X.: CEDNet: A Cascade Encoder-Decoder Network for Dense Prediction. In: Pattern Recognition, 2302.06052 (2023)
27. Liu, Q., Li, X., He, Z., Fan, N., Yuan, D., Wang, H.: Learning deep multi-level similarity for thermal infrared object tracking. IEEE Trans. Multimedia, **23**, 2114-2126 (2020)
28. Song, Y., Ma, C., Wu, X., Gong, L., Bao, L., Zuo, W., Yang, M. H.: Vital: Visual tracking via adversarial learning. In: CVPR, 8990-8999 (2018)
29. Li, F., Tian, C., Zuo, W., Zhang, L., Yang, M. H.: Learning spatial-temporal regularized correlation filters for visual tracking. In: CVPR, 4904-4913 (2018)
30. Nam, H., Han, B.: Learning multi-domain convolutional neural networks for visual tracking. In: CVPR, 4293-4302 (2016)
31. Gao, X., Tang, Z., Deng, Y., Hu, S., Zhao, H., Zhou, G.: HSSNet: A end-to-end network for detecting tiny targets of apple leaf diseases in complex backgrounds. Plants, **12**, 2806 (2023)
32. Danelljan, M., Bhat, G., Khan, F. S., Felsberg, M.: Atom: Accurate tracking by overlap maximization. In: ICCV, 4660-4669 (2019)
33. Zhang, L., Gonzalez-Garcia, A., Van De Weijer, J., et al.: Synthetic data generation for end-to-end thermal infrared tracking. IEEE Trans. Image Process., **28**, 1837-1850 (2018)
34. Danelljan, M., Bhat, G., Shahbaz Khan, F., Felsberg, M.: Eco: Efficient convolution operators for tracking. In: CVPR, 6638-6646 (2017)
35. Bertinetto, L., Valmadre, J., Golodetz, S., Miksik, O., Torr, P. H.: Staple: Complementary learners for real-time tracking. In: CVPR, 1401-1409 (2016)
36. Wang, N., Zhou, W., Tian, Q., Hong, R., Wang, M., Li, H.: Multi-cue correlation filters for robust visual tracking. In: CVPR, 4844-4853 (2018)

37. M. Danelljan, G. Häger, F. S. Khan, and M. Felsberg.: Accurate scale estimation for robust visual tracking. In: Proc. Brit. Mach. Vis. Conf, 1–5 (2014)
38. Zuo, W.; Wu, X.; Lin, L.; Zhang, L.; Yang, M.-H.: Learning support correlation filters for visual tracking. In: IEEE Trans. Pattern Anal, 1158–1172 (2018)
39. Ma C, Huang J B, Yang X, et al.: Hierarchical convolutional features for visual tracking. In: ICCV, 3074-3082 (2015)